\documentclass[sigconf, screen]{acmart}

\newcommand{\ie}{{\emph{i.e.}}, }
\newcommand{\etal}{{\it {et al}.} }

\usepackage{subfigure}

\usepackage{xcolor}
\usepackage{tabularx} 
\usepackage{multirow}
\usepackage{tabu}
\usepackage{pbox}
\usepackage{graphicx}
{\begin{list}               
    {$\bullet$ \hfill}{
        \setlength{\leftmargin}{\parindent}
        \setlength{\parsep}{0.04\baselineskip}
        \setlength{\itemsep}{0.5\parsep}
        \setlength{\labelwidth}{\leftmargin}
        \setlength{\labelsep}{0em}}
    }
{\end{list}}

\providecommand{\eref}[1]{Eq. \eqref{#1}}  
\providecommand{\cref}[1]{Chapter~\ref{#1}}
\providecommand{\sref}[1]{Section~\ref{#1}}
\providecommand{\fref}[1]{Figure~\ref{#1}}
\providecommand{\tref}[1]{Table~\ref{#1}}

\providecommand{\bignorm}[1]{\big\lVert#1\big\rVert}
\providecommand{\norm}[1]{\lVert#1\rVert}

\renewcommand{\vec}[1]{\ensuremath{\boldsymbol{#1}}}
\providecommand{\mat}[1]{\ensuremath{\boldsymbol{#1}}}

\providecommand{\calC}{\mathcal{C}}
\providecommand{\calD}{\mathcal{D}}

\providecommand{\calL}{\mathcal{L}}

\providecommand{\calP}{\mathcal{P}}

\providecommand{\mF}{\mat{F}}

\providecommand{\mI}{\mat{I}}

\providecommand{\mM}{\mat{M}}

\providecommand{\vp}{\vec{p}}

\providecommand{\vy}{\vec{y}}

\copyrightyear{2023}
\acmYear{2023}
\setcopyright{licensedothergov}\acmConference[MM '23]{Proceedings of the 31st ACM International Conference on Multimedia}{October 29-November 3, 2023}{Ottawa, ON, Canada}
\acmBooktitle{Proceedings of the 31st ACM International Conference on Multimedia (MM '23), October 29-November 3, 2023, Ottawa, ON, Canada}
\acmPrice{15.00}
\acmDOI{10.1145/3581783.3612450}
\acmISBN{979-8-4007-0108-5/23/10}

\begin{document}

\title[FDCNet: Feature Drift Compensation Network for Class-Incremental Weakly Supervised Object Localization]{FDCNet: Feature Drift Compensation Network for Class-Incremental Weakly Supervised Object Localization}

\author{Sejin Park}
\authornote{Both authors contributed equally to this research.}
\affiliation{
 \institution{Seoul National University of Science and Technology}
 \city{Seoul}
 \country{Republic of Korea}}
\email{sejin54852@seoultech.ac.kr}

\author{Taehyung Lee}
\authornotemark[1]
\affiliation{
 \institution{Seoul National University of Science and Technology}
 \city{Seoul}
 \country{Republic of Korea}}
\email{22510147@seoultech.ac.kr}

\author{Yeejin Lee}
\affiliation{
 \institution{Seoul National University of Science and Technology}
 \city{Seoul}
 \country{Republic of Korea}}
\email{yeejinlee@seoultech.ac.kr}

\author{Byeongkeun Kang}
\authornote{Corresponding author.}
\affiliation{
 \institution{Seoul National University of Science and Technology}
 \city{Seoul}
 \country{Republic of Korea}}
\email{byeongkeun.kang@seoultech.ac.kr}

\renewcommand{\shortauthors}{Sejin Park, Taehyung Lee, Yeejin Lee, and Byeongkeun Kang}

\begin{abstract}
This work addresses the task of class-incremental weakly supervised object localization (CI-WSOL). The goal is to incrementally learn object localization for novel classes using only image-level annotations while retaining the ability to localize previously learned classes. This task is important because annotating bounding boxes for every new incoming data is expensive, although object localization is crucial in various applications. To the best of our knowledge, we are the first to address this task. Thus, we first present a strong baseline method for CI-WSOL by adapting the strategies of class-incremental classifiers to mitigate catastrophic forgetting. These strategies include applying knowledge distillation, maintaining a small data set from previous tasks, and using cosine normalization. We then propose the feature drift compensation network to compensate for the effects of feature drifts on class scores and localization maps. Since updating network parameters to learn new tasks causes feature drifts, compensating for the final outputs is necessary. Finally, we evaluate our proposed method by conducting experiments on two publicly available datasets (ImageNet-100 and CUB-200). The experimental results demonstrate that the proposed method outperforms other baseline methods.  
\end{abstract}



\maketitle

\section{Introduction}
Weakly supervised object localization (WSOL) is an important research topic because it enables the learning of object localization without requiring expensive bounding box annotations. Instead of relying on bounding boxes, it learns object localization using only image-level class labels, which are easier and less time-consuming to obtain. Naturally, WSOL is more practical than fully supervised approaches, especially when dealing with numerous object categories in the real world. Consequently, many researchers have investigated various methods to achieve accurate object localization by training networks only with image-level class annotations~\cite{Wu2022BAS, ACMMM22XuProxy}. 

\begin{figure}[t]
\centering
\vspace{2mm}
    \includegraphics[width=0.47\textwidth]{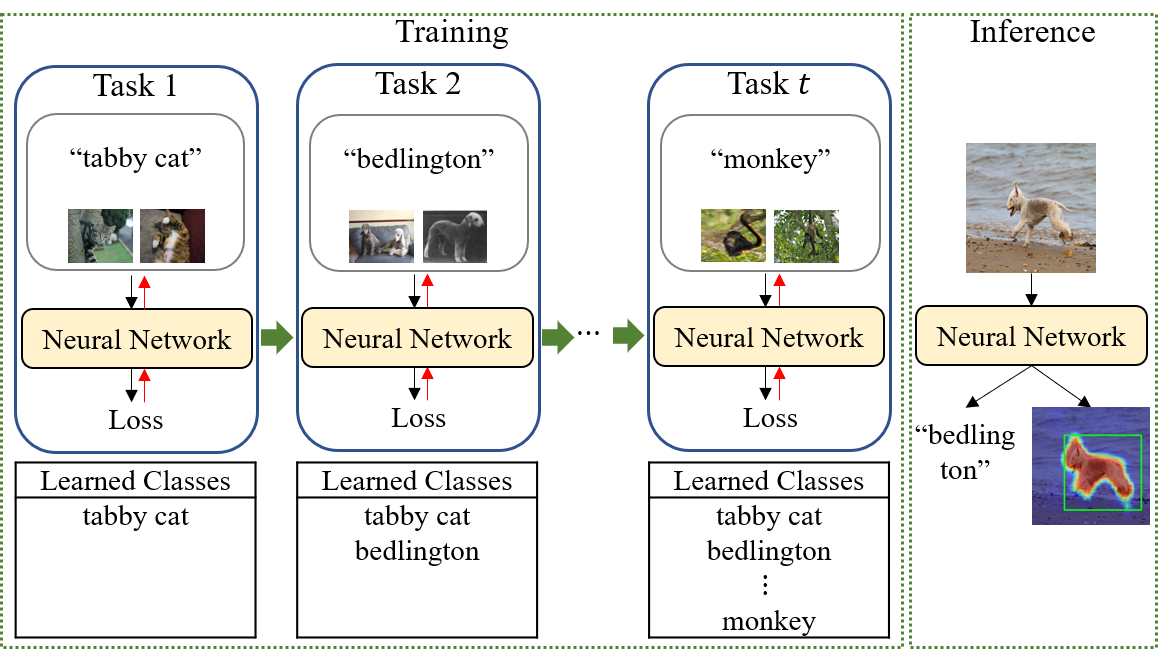}
\vspace{-2mm}
    \caption{Illustration of class-incremental WSOL.}
\label{fig:teaser}
\end{figure}

Although WSOL methods reduce annotation efforts for the training of an object localization network, they have limitations in the object categories that they can localize. Since they learn the localization of objects using a training dataset with predefined categories, they can only localize objects in the predefined classes. This is a significant limitation because the real world contains a vast number of object categories, and new object categories may emerge over time. Hence, it is essential in various applications to be able to learn additional object categories over time without having to re-train all the object classes from scratch. Therefore, this paper addresses the novel problem that aims to incrementally learn object localization for new classes using only image-level class annotations over time while retaining knowledge about previously seen categories (\ie {\it class-incremental WSOL}), as illustrated in~\fref{fig:teaser}.

As class-incremental WSOL (CI-WSOL) enables expandable and progressive object localization, it has many practical applications in various areas such as robotics, autonomous driving, and surveillance. For example, it enables a robot to continuously learn the localization of additional object categories without bounding box annotations. Consequently, it makes the robot easier to perform new tasks and interact with its environment. Moreover, an autonomous vehicle can employ CI-WSOL to adapt to newly introduced electric scooters or traffic signs, improving safety. 

\begin{figure}[!t] 
\centering
\centerline{\includegraphics[scale=0.42]{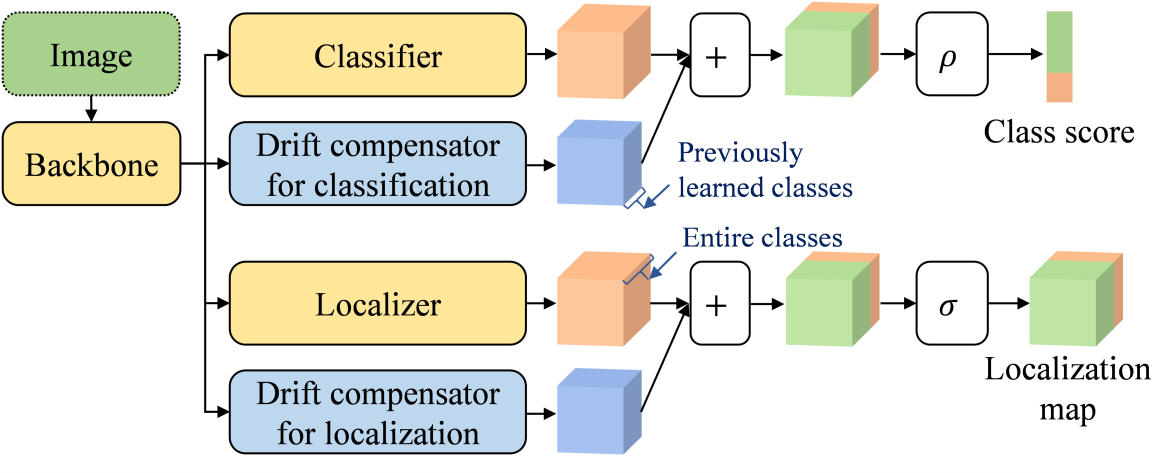}}
\vspace{-2mm}
\caption{Proposed framework during inference.}
\label{fig:framework_inference}
\end{figure}

The contributions of this paper are as follows: (1) We present, to the best of our knowledge, the first class-incremental WSOL method that incrementally learns the localization of novel object categories using only image-level class labels over time. (2) We propose a strong baseline method that employs four knowledge distillation losses, maintains a small subset of previously learned data, and uses cosine normalization, to mitigate catastrophic forgetting. These distillation losses are computed not only using class scores but also localization maps and feature maps. (3) To effectively maintain knowledge about previously seen categories while learning new object classes, we propose to compensate for the effects of feature drifts on both classification and localization using neural networks (see~\fref{fig:framework_inference}). (4) We demonstrate the effectiveness of the proposed method by training CNNs initially on a base set of object classes in a dataset and then incrementally on the remaining categories.

\section{Related Work}
\textbf{Weakly Supervised Object Localization.}
Oquab \etal presented one of the earliest works on WSOL~\cite{Oquab2015WSOL}. They converted the fully connected layers in an image classification network into convolutional layers to preserve spatial dimensions for localization. Additionally, they employed max-pooling to search for object locations using only image-level class annotations. In another early work, Zhou \etal introduced the class activation map (CAM) for localization, which is obtained by a linear combination of feature maps~\cite{Zhou2016CAM}. Since then, researchers have proposed various approaches to improve the accuracy of WSOL~\cite{ACMMM21ShaoImproving, ACMMM21ChenE2Net}. 

Since WSOL methods are trained using only image-level class labels without detailed bounding box annotations, they tend to localize only the most discriminative region rather than the whole object. To overcome this limitation, researchers have proposed diverse methods, such as erasing discriminative regions~\cite{Singh2017HideNSeek, Zhang2018ACoL}, using pseudo-labels for training localizers~\cite{Zhang2018SPG, Zhang2020PseudoBBOX}, and combining multiple feature maps~\cite{Xue2019DANet, Wei2021Shallow, Wu2023multipleFeature}. Recently, Wu \etal proposed the background activation suppression (BAS) method that uses foreground/background activation values directly along with cross-entropy losses~\cite{Wu2022BAS}. This method localizes the whole object regions better by suppressing the activation magnitudes of the background area. Because of its high accuracy, we propose a class-incremental WSOL method by extending the BAS method~\cite{Wu2022BAS}. 

To the best of our knowledge, all previous WSOL methods assume that all training data are given before training begins. To overcome this constraint, we propose a WSOL method that incrementally learns novel object classes using additional data that appear over time while retaining previously learned knowledge.

\textbf{Class-Incremental Learning.}
Li \etal presented one of the earliest works that addressed the problem of catastrophic forgetting in deep neural networks~\cite{Li2016LwF,Li2018LwF}. They proposed to preserve performance on previously learned tasks using knowledge distillation loss~\cite{Hinton2014}, which was computed using the network's outputs for images in the new task. Rebuffi \etal introduced a neural network for class-incremental image classification~\cite{Rebuffi2017iCaRL}. They proposed to use a subset of previous training data (\ie exemplar set) and knowledge distillation losses. Additionally, their method classifies test data using the nearest mean of exemplars. Since then, researchers have proposed diverse approaches for class-incremental learning~\cite{ACMMM21ZhouCo}. These methods include employing parameter regularization~\cite{kirkpatrick2017}, investigating better knowledge distillation~\cite{Dhar2019LwM}, using an exemplar set~\cite{Castro2018End}, and correcting biases~\cite{Hou2019Rebalancing}.

Forgetting previously learned knowledge occurs when adding and updating network parameters to learn new tasks. To learn novel classes without adding new weights, Yu \etal proposed using embedding networks for incremental learning~\cite{Yu2020SemanticDriftCompensation}. They also introduced a method to estimate the semantic drift of features, which is the difference between the true class mean feature vector and the estimated class mean feature vector, and to compensate for it. Iscen \etal proposed preserving feature vectors of images in the previous tasks rather than the images themselves, to reduce memory consumption for an exemplar set~\cite{Iscen2020MemoryEfficient}. Since the network's parameters are updated to learn a new task over time, they also proposed the feature adaptation network that translates the previously stored features into the updated feature space. Wu \etal hypothesized that a strong backbone trained using a large number of base classes is available, and the backbone can extract meaningful representations even for novel classes~\cite{Wu2022Strong}. Then, they replicated and fine-tuned only the last layers for novel classes. Finally, they combined the networks' outputs for base classes and the replicated networks' outputs for novel classes by their score fusion method.

In this paper, we first propose a strong baseline method for class-incremental WSOL, which we believe is the first of its kind. We then introduce the feature drift compensation network for CI-WSOL, which compensates for the effects of feature drifts on class scores and localization maps using neural networks. Whereas Yu \etal compensated for the drifts of class mean feature vectors by interpolating sparse vectors for the nearest class mean classifier~\cite{Yu2020SemanticDriftCompensation}, we train neural networks to compensate for class scores and activation maps using the feature maps near the outputs. Whereas Iscen \etal transformed the stored features into a new feature space to deal with changes in network parameters~\cite{Iscen2020MemoryEfficient}, we estimate the effects of changes in feature spaces on the final outputs. Lastly, whereas Wu \etal employed a fixed backbone and replicated classifiers to learn new tasks~\cite{Wu2022Strong}, we allow all parameters to be updated to better learn novel classes and do not require replicated classifiers.

\section{Class-Incremental WSOL}\label{sec:CI-WSOL}
We first define the class-incremental WSOL problem that aims to class-incrementally learn object localization given only images and their image-level class labels. Then, we introduce a baseline WSOL network and extend this to propose a strong baseline method for class-incremental learning. 

\subsection{Problem Formulation}
Initially, a training dataset $\calD^1$ is given where $\calD^1$ contains the images and their image-level class labels of $\calC^1$ base classes. Then, an object localization neural network $f^1$ is trained using the dataset $\calD^1$. It is called the first task ($t=1$). Later, at task $t=2$, an additional dataset $\calD^2$ of novel classes $\calC^2$ is provided while the previous dataset $\calD^1$ is no longer available. We assume that the object classes between two distinct tasks are disjoint (\ie $\calC^i \cap \calC^j = \emptyset \text{ if } i \neq j$). In this circumstance, we aim to obtain the new network $f^2$ that can localize both $\calC^1$ and $\calC^2$ object classes using the previously trained network $f^1$ and the new dataset $\calD^2$. Similarly, at a subsequent task $t>2$, a new network $f^t$ is trained to accurately localize accumulated object classes $\calC^t_{acc} = \cup_{i=1}^{t} \calC^i$ given the previous network $f^{t-1}$ and a new dataset $\calD^t$ of novel classes $\calC^t$.

In summary, at each task $t>1$, we are given a new dataset $\calD^t$ containing novel objects $\calC^t$ and the previously trained network $f^{t-1}$ from the previous task. Then, when the training finishes at this task, the new network $f^t$ is expected to accurately localize the objects of both the novel classes $\calC^t$ in the new dataset and the previously learned classes $\calC^{t-1}_{acc}$ until the previous task. Therefore, the localization network should be able to incrementally learn new objects while preserving its knowledge for already known classes. 

\subsection{WSOL Network}
Given an image, a WSOL network aims to predict both the object class and its bounding box. A typical WSOL network $f_w$ consists of a feature extractor $f_{b}$, a localizer $f_{l}$, and a classifier $f_{c}$. The feature extractor takes an input image $\mI$ and extracts a shared feature map $\mF$. Then, both $f_{l}$ and $f_{c}$ take $\mF$, and they produce a class activation map $\mM^{\text{CAM}} \in  \mathbb{R}^{K \times N_1 \times N_1}$ and a class score map $\mM^{cls} \in \mathbb{R}^{K \times N_2 \times N_2}$, respectively, where $K$ denotes the number of classes; $N_1$ and $N_2$ represent the spatial dimensions of $\mM^{\text{CAM}}$ and $\mM^{cls}$, respectively. Since $\mM^{\text{CAM}}_{k,i,j}$ represents the probability of having the $k$-th object at ($i,j$) location, a foreground mask $\mM^{fg}$ for an object class is obtained by slicing the corresponding channel of $\mM^{\text{CAM}}$. Then, the bounding box for localization is typically calculated by finding the surrounding box of the largest foreground blob. We abbreviate the sequential process of $f_{b}$ and $f_{c}$ by $f_{b,c}$, and that of $f_{b}$ and $f_{l}$ by $f_{b,l}$, respectively (\ie $f_{b,c}(\cdot)=f_c(f_b(\cdot))$ and $f_{b,l}(\cdot)=f_l(f_b(\cdot))$).

Since only image-level class labels are given, the network is typically trained using a loss function containing two cross-entropy losses. One is typical image classification loss $\calL_{cls}$ that is computed using the class score map $\mM^{cls}$ from the classifier and the one-hot encoded class label $\vy$. Hence, this loss depends on only the feature extractor and the classifier of the network. 
\begin{equation}
\calL_{cls}(\mI, \vy) := - \sum_{i=1}^{K} \vy_i  \ln\Big( \rho(f_{b,c}(\mI))_{i} \Big) 
\label{eq:loss_class}
\end{equation}
where $i$ is an index for object classes, and $K$ is the total number of learned classes; $\rho$ denotes consecutive processing of global average pooling and applying a softmax function. 

To train the localizer along with other components using the image-level class labels, the other cross-entropy loss is computed using both the foreground mask and class score map. As $\mM^{fg}$ aims to highlight the region containing the corresponding object, the element-wise multiplication of $\mM^{fg}$ and $\mM^{cls}$ should produce a high classification score. Accordingly, the foreground classification loss $\calL_{cls\text{-}fg}$ is computed as follows: 
\begin{equation}
\calL_{cls\text{-}fg}(\mI, \vy) := -\sum_{i=1}^{K} \vy_i \ln \Bigl( \rho(f_{b,c}(\mI) \cdot \sigma_{\downarrow}(f_{b,l}(\mI)_{i}) )_{i} \Bigr) 
\label{eq:loss_class_fg}
\end{equation}
where $\sigma_{\downarrow}$ represents sequential processing of applying a sigmoid function to each element and downsampling to match the dimension of $\mM^{cls}$. ($\cdot$) in~\eref{eq:loss_class_fg} denotes channel-wise multiplication\footnote{$\sigma_{\downarrow}(f_{b,l}(\mI)_{i})$ is multiplied to each channel of $f_{b,c}(\mI)$.}.

Since the network trained using only the two cross-entropy losses tends to highlight only the most discriminative region, we also employ the background activation suppression loss and area constraint loss in~\cite{Wu2022BAS}, to localize an entire object. The background activation suppression loss $\calL_{bas}$ is computed as follows:
\begin{equation}
\begin{split}
	\calL_{bas}(\mI, y) := \frac{s_{bg}(\mI, y)}{s_{all}(\mI, y) + \epsilon} 
\end{split}
\label{eq:loss_BAS}
\end{equation}
where $s_{all}$ and $s_{bg}$ compute the average scores over the entire image and over the background region, respectively. $y$ and $\epsilon$ denote the ground-truth class and a small value to avoid dividing by zero, respectively. Specifically, they are defined as follows:
\begin{equation}
\begin{split}
	s_{all}(\mI, y) := \text{GAP}&\big(f_{b,c}(\mI)_y\big),  \\
	s_{bg}(\mI, y)  := \text{GAP}&\Big(f_c \big(f_b(\mI) \cdot (1 - \sigma(f_{b,l}(\mI))_y) \big)_y \Big) 
\end{split}
\end{equation}
where $\text{GAP}$ and $\sigma$ represent global average pooling and applying a sigmoid function to each element, respectively. 

The area constraint loss $\calL_{ac}$ is computed as follows:
\begin{equation}
	\calL_{ac}(\mI, y) := \frac{1}{N} \sum_i \sum_j \sigma\big(f_{b,l}(\mI) \big)_{y,i,j} 
\label{eq:loss_AC}
\end{equation}
where $N$ is the total number of elements in $\mM^{fg}$. 

In summary, the total loss $\calL_{wsol}$ for the baseline WSOL network is as follows:
\begin{equation}
\begin{split}
\calL_{wsol} := \calL_{cls} + \alpha_1 \calL_{cls\text{-}fg} + \alpha_2 \calL_{bas} + \alpha_3 \calL_{ac}
\end{split}
\label{eq:total_WSOL}
\end{equation}
where $\alpha_1$, $\alpha_2$, and $\alpha_3$ are hyperparameters to balance the losses. 

For more details about the baseline WSOL method, please refer to BAS~\cite{Wu2022BAS} and~\fref{fig_framework}. 

\subsection{Strong Baseline for Class-Incremental WSOL} \label{sec:strong_baseline}
Catastrophic forgetting is one of the main challenges in incremental learning~\cite{McCloskey1989Catastrophic, Goodfellow2013Catastrophic}. To alleviate catastrophic forgetting in CI-WSOL, we employ knowledge distillation losses, maintain a small data set among previous data (\ie exemplar set $\calP$), and use cosine normalization. Different from the methods for classification~\cite{Li2018LwF}, we compute knowledge distillation losses using not only classification scores and feature maps but also localization maps. 

\noindent\textbf{Knowledge distillation.}
It was introduced in LwF~\cite{Li2018LwF} to alleviate catastrophic forgetting during training a class-incremental classifier. It encourages the outputs of the updated network to be the same as those of the previous model. These losses are in the form of either a modified cross-entropy loss~\cite{Li2018LwF} or Kullback-Leibler (KL) divergence~\cite{1951KLDivergence}. Since the effects of the two are almost the same, we employ KL divergence-based knowledge distillation losses. 

A typical knowledge distillation loss $\calL_{kd\text{-}cls}$ for a classifier uses the logits of the two networks. Because $f_{b,c}^t$ and $f_{b,c}^{t-1}$ predict the scores for $\calC^t_{acc}$ and $\calC^{t-1}_{acc}$ classes, respectively, the scores for only $\calC^{t-1}_{acc}$ classes are utilized to compute this loss. 
\begin{equation}
\begin{split}
	& \calL_{kd\text{-}cls}(\mI) := D_{\text{KL}}\Big(\rho \big(f_{b,c}^{t-1}(\mI) \big) || \rho \big(f_{b,c}^t(\mI)\big)_{1:|\calC^{t-1}_{acc}|} \Big)
\end{split}
\label{eq:loss_KD_logit}
\end{equation}
where $D_{\text{KL}}(\cdot||\cdot)$ represents the KL-divergence between two inputs. $|\calC^{t-1}_{acc}|$ denotes the number of classes in the set $\calC^{t-1}_{acc}$.

Different from the networks for classification, a WSOL network needs an additional constraint to less forget localization. Hence, we propose an additional distillation loss using foreground masks. Whereas class probabilities are relative to those for other classes, the foreground for a class is estimated independently from those for other classes by applying a sigmoid function to each element. Therefore, this distillation loss $\calL_{kd\text{-}loc}$ is computed using only the foreground map for the ground-truth class as follows:
\begin{equation}
\begin{split}
	\calL_{kd\text{-}loc}(\mI, y) := \bignorm{\sigma(f_{b,l}^{t}(\mI))_y - \sigma(f_{b,l}^{t-1}(\mI))_y}_2.
\end{split}
\label{eq:loss_KD_loc}
\end{equation}
Because $f_{b,l}^{t-1}$ predicts the foreground for only $\calC^{t-1}_{acc}$ classes, this loss is computed using only the data in the exemplar set $\calP$. 

Moreover, a distillation loss using feature descriptors was proposed as an additional constraint in classification~\cite{Less2018Jung}. Hence, we also employ two distillation losses using feature maps to encourage the feature maps of $f^t$ to be close to those of $f^{t-1}$. The two losses use the feature maps that are one layer prior to the class score map and the localization map, respectively. These distillation losses $\calL_{kd\text{-}feat\text{-}cls}$ and $\calL_{kd\text{-}feat\text{-}loc}$ are computed using cosine similarities as follows:
\begin{equation}
\begin{split}
&\calL_{kd\text{-}feat\text{-}cls}(\mI) := 1 - s_c\Big(\hat{f}^{t}_{b,c}(\mI), \hat{f}^{t-1}_{b,c}(\mI)\Big), \\
&\calL_{kd\text{-}feat\text{-}loc}(\mI) := 1 - s_c\Big(\hat{f}^{t}_{b,l}(\mI), \hat{f}^{t-1}_{b,l}(\mI)\Big), 
\end{split}
\label{eq:loss_KD_feature}
\end{equation}
where $s_c(\cdot, \cdot)$ denotes the function that computes the average of the cosine similarities between the two vectors at the same location in the two input matrices; $\hat{f}_{b,c}$ and $\hat{f}_{b,l}$ represent the networks that exclude the last layer from $f_{b,c}$ and $f_{b,l}$, respectively.

\begin{figure*}[!t] 
\centering
\begin{minipage}{0.98\linewidth}
\centerline{\includegraphics[scale=0.57]{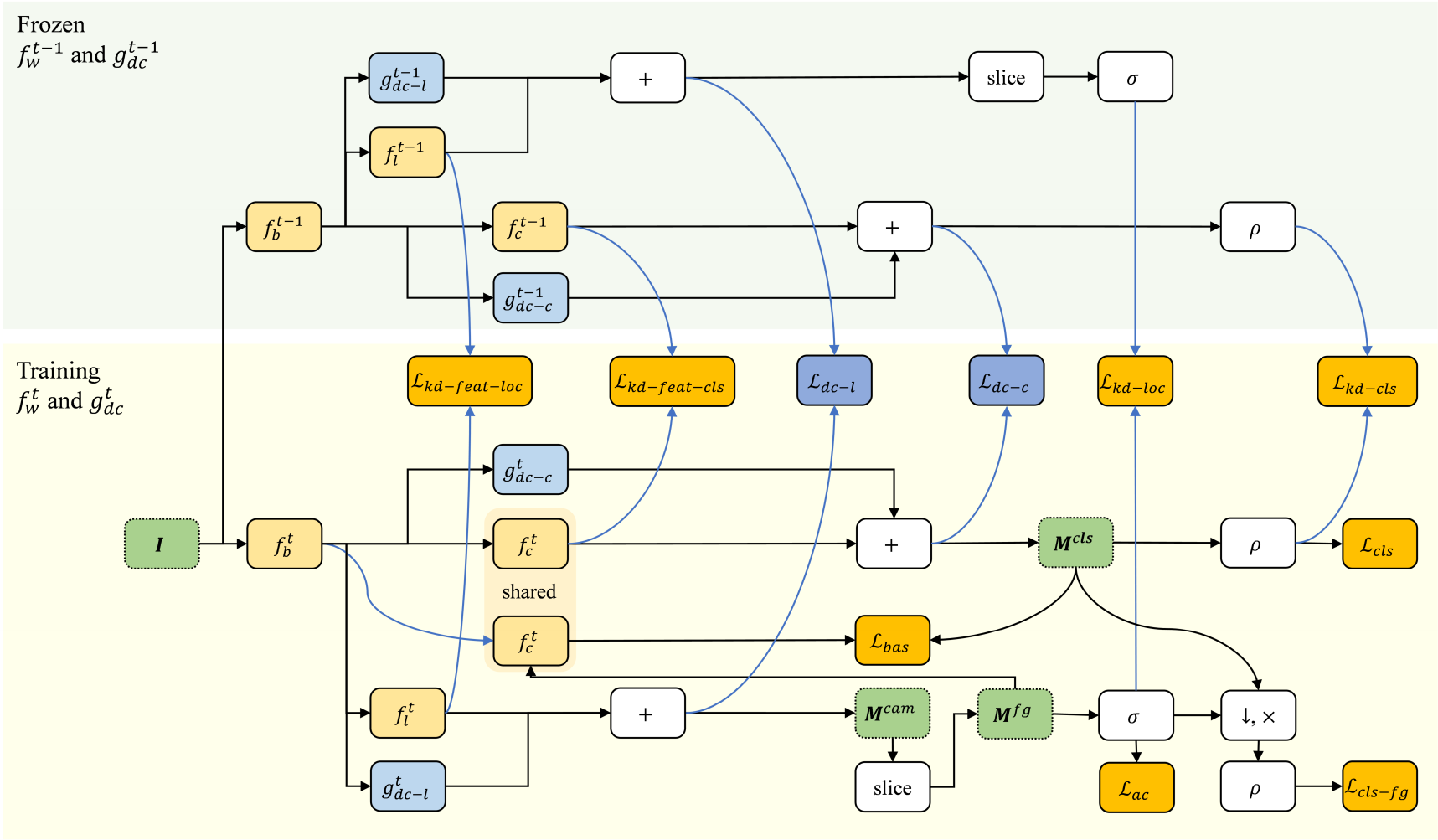}}
\end{minipage}
\caption{Proposed framework during training. Bright and dark yellow boxes denote the modules ($f_b$, $f_c$, and $f_l$) in the WSOL network, and the losses to train them, respectively. Bright and dark blue boxes represent the FDC modules ($g_{dc\text{-}c}$ and $g_{dc\text{-}l}$) and the losses to train them, respectively. Green and white boxes denote input/outputs and non-learnable operations, respectively.}
\label{fig_framework}
\end{figure*}

These distillation losses are used to train the CI-WSOL network in addition to $\calL_{wsol}$ in~\eref{eq:total_WSOL} (see~\fref{fig_framework}). Therefore, the total loss is as follows: 
\begin{equation}
\begin{split}
\calL_{ci\text{-}wsol}:= \alpha_4 & \calL_{kd\text{-}cls} + \alpha_5 \calL_{kd\text{-}loc} + \alpha_6 \calL_{kd\text{-}feat\text{-}cls} \\
				+ \alpha_7 & \calL_{kd\text{-}feat\text{-}loc} + \calL_{wsol.}
\end{split}
\label{eq:loss_CI_WSOL}
\end{equation}
where $\alpha_4$, $\alpha_5$, $\alpha_6$, and $\alpha_7$ are hyperparameters to balance the losses. 

\noindent\textbf{Maintaining data of previous tasks.}
In addition to knowledge distillation, a subset of the previously learned data is retained and utilized to mitigate catastrophic forgetting~\cite{Rebuffi2017iCaRL}. Most methods select a collection of representative examples from the data at the previous tasks, which is called an \textit{exemplar} set, by using the herding algorithm~\cite{Welling2009Herding}. Then, the exemplar set is used along with the newly provided dataset during training for the current task. Following previous works~\cite{Rebuffi2017iCaRL}, we also preserve a subset of representative data from previous tasks using the herding algorithm~\cite{Welling2009Herding}. 

At the $t$-th task ($t>1$), the network is trained using the new training dataset $\calD^t$ containing novel classes $\calC^t$ and the exemplar set $\calP$ with previously learned classes $\calC^{t-1}_{acc}=\cup_{i=1}^{t-1} \calC^i$. After training for the $t$-th task, the exemplar set is updated by excluding less representative data from $\calP$ and then by adding the newly selected data from $\calD^t$ to $\calP$. The memory requirement for $\calP$ is constrained by fixing the total number of examples in $\calP$ during the entire training process by first excluding and then adding. 

To select representative examples, the mean feature descriptor for each class is utilized. Whereas the networks for classification use the features near the final output of a classifier, we need to consider both classification and localization performances to select representative examples. Therefore, we use the shared feature maps (\ie outputs of $f_b$) to estimate representativeness.

\noindent\textbf{Cosine normalization.}
While an exemplar set is used, the data for the current task are more prevalent than those for the previous tasks. Accordingly, the network tends to produce higher scores for the lastly learned classes $\calC^t$ than those for the other classes $\calC^{t-1}_{acc}$. Therefore, cosine normalization is employed to balance magnitudes across all the classes following~\cite{Hou2019Rebalancing}. 

Specifically, it is applied to the last layer in the classifier $f_c$. Hence, a class score at a pixel is computed by first multiplying two $\ell_2$-normalized vectors for the class and then scaling using a learnable scalar. The two vectors before normalization are a weight vector for the class and a feature vector in the feature map from the previous layer (\ie outputs of $\hat{f}_{b,c}$). Since the output of $\hat{f}_{b,c}$ is a feature map rather than a vector, it is applied at all the locations to obtain a class score map $\mM^{cls}$.

\section{Feature Drift Compensation} \label{sec:fdc}
While strong knowledge distillation and regularization can mitigate catastrophic forgetting, they limit the capability of the network for learning new tasks because of the constraint on the parameters in the network. Therefore, we propose a novel method that can mitigate catastrophic forgetting while allowing flexibility to the network's parameters to learn new tasks well. 

During training on a new task, parameters in the WSOL network are updated. Accordingly, extracted features using the updated network are drifted from those using the previous network. Therefore, we propose two modules that estimate the effects of the drifts on classification scores and localization maps, and compensate for them (\ie one for the classifier and the other for the localizer). Hence, the proposed module enables the WSOL network to produce comparable outputs for the previously learned classes while evolving over time to learn new tasks. We call the modules feature drift compensation (FDC) network.

\subsection{Overview} \label{sec:fdc_overview}
Feature drift compensation (FDC) network is trained after training the WSOL network on each task. We first describe the training of the FDC network for the initial three tasks and then extend it to subsequent tasks. At task $t=1$, a WSOL network $f^1_w$ is trained by back-propagating $\calL_{wsol}$ using the training dataset $\calD^1$. After the training completes, we create an exemplar set $\calP^1$ that contains a subset of $\calD^1$ by using the herding algorithm~\cite{Welling2009Herding}. At task $t=2$, $f^2_w$ is created by copying $f^1_w$ and by expanding the last layers of both the classifier and the localizer to learn novel classes. The network is then trained by back-propagating $\calL_{ci\text{-}wsol}$ using both the new dataset $\calD^2$ and the exemplar set $\calP^1$. $f^1_w$ is frozen and utilized during training to compute the distillation losses and to train a FDC network. After the training finishes, the exemplar set $\calP^2$ is updated by excluding less-representative examples from $\calP^1$ and by adding representative samples from $\calD^2$.

During training $f^2_w$, the parameters in $f^2_w$ are drifted from those in $f^1_w$. The change of the parameters causes the drift of feature maps, which further affects classification and localization performances. Therefore, we train a FDC network $g^2_{dc}$ to map the feature maps from $f^2_w$ to the differences between the outputs of $f^1_w$ and those of $f^2_w$. The network $g^2_{dc}$ is trained by back-propagating $\calL_{dc}$ which is introduced in~\sref{sec:trainingFDC}. 

At task $t=3$, $f^3_w$ is created by expanding $f^2_w$ by the number of novel classes $|\calC^3|$. Then, it is initialized using $f^2_w$ and then trained by back-propagating $\calL_{ci\text{-}wsol}$ using both $\calD^3$ and $\calP^2$. After the training completes, the updated exemplar set $\calP^3$ is obtained. Lastly, a FDC network $g^3_{dc}$ is trained using $\calL_{dc}$ to map the feature maps of $f^3_w$ to the discrepancy between the outputs of $f^3_w$ and the summation of those of $f^2_w$ and $g^2_{dc}$. 

The procedure at $t=3$ is repeated for subsequent tasks $t>3$. Note that a FDC network compensates for only the previously learned classes $\calC^{t-1}_{acc}$ to preserve knowledge. It is apparent since target scores can be computed for only $\calC^{t-1}_{acc}$, using $f^{t-1}_w$ and $g^{t-1}_{dc}$.


\subsection{FDC Network}
The FDC network consists of two modules where one for classification and the other for localization. The former $g^t_{dc\text{-}c}$ maps the feature maps from an intermediate layer in $f^t_{b,c}$ to the difference between the class scores from $f^t_{b,c}$ and the target class scores. The latter $g^t_{dc\text{-}l}$ maps the feature maps from $\hat{f}^t_{b,l}$ to the discrepancy between the localization map from $f^t_{b,l}$ and the target localization map. The targets for the former and the latter are the summation of $f^{t-1}_{b,c}$ and $g^{t-1}_{dc\text{-}c}$ and that of $f^{t-1}_{b,l}$ and $g^{t-1}_{dc\text{-}l}$, respectively, for $t>2$. When $t=2$, they are the outputs of $f^1_{b,c}$ and $f^1_{b,l}$, respectively, as explained in~\sref{sec:fdc_overview}. 


After training the FDC network at the $t$-th task, a class prediction is made for an image by adding the scores from $f^t_{b,c}$ and $g^{t}_{dc\text{-}c}$ and by finding the argument of the maxima after global average pooling. The localization map is obtained by adding the maps from $f^t_{b,l}$ and $g^{t}_{dc\text{-}l}$ and by applying a sigmoid function at each pixel and each channel. As mentioned earlier, the summation of the two corresponding maps is only for the $\calC^{t-1}_{acc}$ classes while the scores for the $\calC^t$ classes are directly from the final network $f^t_w$. It is evident since the compensation modules estimate for only previously learned classes. 

Therefore, the final class probability vector $\vp$ and localization map $\mM^{loc}$ are computed as follows:
\begin{equation}
\begin{split}
&\vp = \rho\Big(\big[f^t_{b,c}(\mI)_{1:|\calC^{t-1}_{acc}|} + g^{t}_{dc\text{-}c}(f^t_b(\mI)); f^t_{b,c}(\mI)_{|\calC^{t-1}_{acc}|+1:|\calC^{t}_{acc}|}\big]\Big), \\
&\mM^{loc} = \sigma\Big(\big[f^t_{b,l}(\mI)_{1:|\calC^{t-1}_{acc}|} + g^{t}_{dc\text{-}l}(f^t_b(\mI)); f^t_{b,l}(\mI)_{|\calC^{t-1}_{acc}|+1:|\calC^{t}_{acc}|}\big]_y \Big) 
\end{split}
\label{eq:feature_drift_compensation_inference}
\end{equation}
where $[\hspace{1mm};\hspace{1mm}]$ denotes concatenating two matrices.

\subsection{Training FDC Network} \label{sec:trainingFDC}
Since the FDC network is to retain knowledge about the previously learned classes by preserving the outputs of the previous networks, we train the network to minimize the discrepancy between the outputs of the previous networks and the current networks. For $t>2$, the outputs of the previous networks (\ie target scores) are the summation of $f^{t-1}_{b,c}$ and $g^{t-1}_{dc\text{-}c}$ and that of $f^{t-1}_{b,l}$ and $g^{t-1}_{dc\text{-}l}$. The outputs of the current networks are the summation of $f^t_{b,c}$ and $g^t_{dc\text{-}c}$ and that of $f^t_{b,l}$ and $g^t_{dc\text{-}l}$. Among the networks, only $g^t_{dc\text{-}c}$ and $g^t_{dc\text{-}l}$ are trained by minimizing the $\ell_2$ distances between the corresponding outputs whereas all others are frozen. 

Therefore, the losses $\calL_{dc\text{-}c}$ and $\calL_{dc\text{-}l}$ for $g^t_{dc\text{-}c}$ and $g^t_{dc\text{-}l}$, respectively, are computed as follows:
\begin{equation}
\begin{split}
&\calL_{dc\text{-}c}(\mI) :=  \norm{\mM^t_{cls} - \mM^{t-1}_{cls}}_2, \\ 
&\calL_{dc\text{-}l}(\mI) := \norm{\mM^t_{cam} - \mM^{t-1}_{cam}}_2, \\ 
\end{split}
\label{eq:feature_drift_compensation_cls}
\end{equation}
where $\mM^t_{cls} = \big[f^t_{b,c}(\mI)_{1:|\calC^{t-1}_{acc}|} + g^{t}_{dc\text{-}c}(\bar{f}^t_{b,c}(\mI))\big]$,
\begin{equation*}
\begin{split}
&\mM^{t-1}_{cls} = \big[f^{t-1}_{b,c}(\mI)_{1:|\calC^{t-2}_{acc}|} + g^{t-1}_{dc\text{-}c}(\bar{f}^{t-1}_{b,c}(\mI)); f^{t-1}_{b,c}(\mI)_{|\calC^{t-2}_{acc}|+1:|\calC^{t-1}_{acc}|}\big], \\
&\mM^t_{cam} = \big[f^t_{b,l}(\mI)_{1:|\calC^{t-1}_{acc}|} + g^{t}_{dc\text{-}l}(f^t_b(\mI))\big], \\
&\mM^{t-1}_{cam} = \big[f^{t-1}_{b,l}(\mI)_{1:|\calC^{t-2}_{acc}|} + g^{t-1}_{dc\text{-}l}(f^{t-1}_b(\mI)); f^{t-1}_{b,l}(\mI)_{|\calC^{t-2}_{acc}|+1:|\calC^{t-1}_{acc}|}\big];
\end{split}
\end{equation*}
$\bar{f}_{b,c}$ denotes the network excluding the last three layers from $f_{b,c}$. When $t=2$, we can assume that the outputs of $g^{t-1}_{dc\text{-}c}$ and $g^{t-1}_{dc\text{-}l}$ are zero matrices, as explained in~\sref{sec:fdc_overview}. 

The total loss $\calL_{dc}$ for the FDC network is the summation of the two losses where each is responsible for compensating class scores and localization maps.
\begin{equation}
\begin{split}
\calL_{dc} :=& \calL_{dc\text{-}c} + \beta \calL_{dc\text{-}l}
\end{split}
\label{eq:loss_total_fdc}
\end{equation}
where $\beta$ is a hyperparameter to balance the losses.

\section{Experiments and Results}

\subsection{Experimental Settings}
\noindent\textbf{Dataset.}
We experiment on two publicly available datasets, ImageNet-100~\cite{ILSVRC15} and CUB-200~\cite{cub_dataset}, to demonstrate incremental learning of object localization using only weak supervision. The ImageNet-100 dataset is a subset of the ILSVRC-2012 dataset~\cite{ILSVRC15} and has been used in class-incremental learning~\cite{Yu2020SemanticDriftCompensation, Iscen2020MemoryEfficient, Rebuffi2017iCaRL}. We selected this dataset because the ILSVRC-2012 dataset has also been used in the WSOL literature~\cite{Wu2022BAS}. This dataset consists of the initial 100 classes of the ILSVRC-2012 dataset~\cite{ILSVRC15}, and contains 129,395 images for training and 5,000 images for testing. The CUB-200 dataset~\cite{cub_dataset} is another popular dataset in WSOL~\cite{Wu2022BAS}. It contains 200 classes and consists of 5,994 images for training and 5,794 images for testing. 

\vspace{1mm}
\noindent\textbf{Incremental learning.}
The proposed network is trained using images of disjointly selected classes for each task. In the first task, images of $\calC^1$ base classes are used for training. Then, it is incrementally trained using images of $\calC^i$ novel classes at each subsequent task. Therefore, the total number of classes that the network learns at the end is $|\calC^1| + (T-1) |\calC^i|$, where $T$ is the total number of tasks.

For the ImageNet-100 dataset~\cite{ILSVRC15}, $|\calC^1|$ and $|\calC^i|$ are 50 and 10 classes, respectively, and for the CUB-200 dataset~\cite{cub_dataset}, they are 100 and 20 classes, respectively. $T$ is 6 for both datasets. 

\vspace{1mm}
\noindent\textbf{Evaluation metric.}
For quantitative comparison, we measure the \textit{average incremental accuracy} ($Acc_{avg}$) and the \textit{last incremental accuracy} ($Acc_{last}$). The former, $Acc_{avg}$, is the average of accuracies that are computed after each task $t$ using the test data containing all the previously learned classes, $\calC^{t}_{acc}$, (\ie $Acc_{avg}=\frac{1}{T} \sum_{t=1}^T Acc_t$, where $Acc_t$ is the accuracy computed after the $t$-th task). The latter, $Acc_{last}$, is the accuracy computed when the entire training is completed (\ie $Acc_{last}=Acc_{T}$).

To compute the accuracy $Acc_t$ after each task, we use the three typical metrics in WSOL~\cite{Wu2022BAS}, which are Top-1, Top-5, and GT-known localization (Loc) accuracies. Therefore, we report a total of six accuracies for each experiment, as shown in Tables~\ref{tbl:result_cub200} and~\ref{tbl:result_imagenet100}.

\vspace{1mm}
\noindent\textbf{Implementation details.}
We experimented with two different backbones: MobileNetV1~\cite{mobilenetv1} and InceptionV3~\cite{inceptionv3}. The classifier $f_{c}$ and the localizer $f_{l}$ consist of five and one convolution layer(s), respectively. Among the five layers in $f_{c}$, the last layer employs cosine normalization as explained in~\sref{sec:strong_baseline}. Both FDC modules $g_{dc\text{-}c}$ and $g_{dc\text{-}l}$ consist of three convolution layers.

For the ImageNet-100 dataset~\cite{ILSVRC15}, we train for 12 epochs at the first task for both backbones, and then for 70 and 40 epochs at each incremental task for MobileNetV1~\cite{mobilenetv1} and InceptionV3~\cite{inceptionv3}, respectively. Regarding the CUB-200 dataset~\cite{cub_dataset}, we train for 50 and 150 epochs at the first and each incremental task, respectively, for MobileNetV1~\cite{mobilenetv1}, and 100 and 200 epochs for InceptionV3~\cite{inceptionv3}.

\begin{table*}[!t]
\centering
\begin{minipage}{0.97\linewidth}
\caption{Quantitative comparison on the CUB-200 dataset~\cite{cub_dataset}.}
\label{tbl:result_cub200}
\centering
\begin{tabular}{ >{\centering}m{0.3\textwidth}| >{\centering}m{0.13\textwidth}| *2{>{\centering}m{0.06\textwidth} >{\centering}m{0.06\textwidth}|} >{\centering}m{0.06\textwidth}  >{\centering\arraybackslash}m{0.06\textwidth} } 
\toprule
\multirow{2}{*}{Method} & \multirow{2}{*}{Backbone} & \multicolumn{2}{c|}{Top-1 Loc} & \multicolumn{2}{c|}{Top-5 Loc} & \multicolumn{2}{c}{GT-known} \\ 
\cline{3-8}
  &   & $Acc_{avg}$ & $Acc_{last}$ &  $Acc_{avg}$ & $Acc_{last}$ &  $Acc_{avg}$ & $Acc_{last}$ \\
\midrule
FOSTER~\cite{Wang2022foster} + BAS~\cite{Wu2022BAS} & & 47.87 & 30.38 & 60.45 & 39.48  &  67.79 &  43.45 \\
AANets~\cite{Liu2021Adaptive} + BAS~\cite{Wu2022BAS} & & 38.79  & 34.59  & 50.45  &  45.27 &  55.56 & 50.72  \\
Rebalancing~\cite{Hou2019Rebalancing} + BAS~\cite{Wu2022BAS} & \multirow{2}{*}{InceptionV3} & 42.37 & 40.43 & 53.89  &  51.23 & 59.40  & 56.97  \\
CCIL~\cite{Mittal2021CCIL} + BAS~\cite{Wu2022BAS} & \multirow{2}{*}{\cite{inceptionv3}}  & 55.43 & 45.85 & 72.69 & 63.31 &  79.19&  72.65\\ 
FDCNet w/o compensation (ours)  &  &  60.34 & 55.69 &  79.75 & 76.27 & 87.12  & 85.20  \\ 
FDCNet (ours)   &     & \textbf{61.78}  &  \textbf{57.80}  & \textbf{81.80} & \textbf{79.38} & \textbf{89.42} & \textbf{89.21} \\ \cline{1-1}\cline{3-8}
Joint training BAS~\cite{Wu2022BAS} &     & \multicolumn{2}{c|}{73.29} & \multicolumn{2}{c|}{86.31} & \multicolumn{2}{c}{92.24} \\   
\midrule
FOSTER~\cite{Wang2022foster} + BAS~\cite{Wu2022BAS} & & 37.42 & 34.25 & 50.45 & 46.80 & 60.06 & 55.64 \\ 
AANets~\cite{Liu2021Adaptive} + BAS~\cite{Wu2022BAS} & & 47.98 & 39.51 & 57.50 & 53.10 & 67.03 & 63.85 \\ 
Rebalancing~\cite{Hou2019Rebalancing} + BAS~\cite{Wu2022BAS} & \multirow{2}{*}{MobileNetV1}& 46.58 & 44.03 & 61.27 & 58.79 & 68.49 & 67.25 \\
CCIL~\cite{Mittal2021CCIL} + BAS~\cite{Wu2022BAS} & \multirow{2}{*}{\cite{mobilenetv1}} & 53.99 & 46.05 & 74.20 & 65.90 &  84.37&  78.37\\ 
FDCNet w/o compensation (ours)  &  & 56.15 & 51.20 & 76.54 & 73.32 & 87.17 & 87.31 \\ 
FDCNet (ours)   &     &  \textbf{57.01} &  \textbf{52.23} & \textbf{76.93} & \textbf{73.98} & \textbf{88.32} & \textbf{87.80} \\ \cline{1-1}\cline{3-8}
Joint training BAS~\cite{Wu2022BAS}  &     & \multicolumn{2}{c|}{69.77} & \multicolumn{2}{c|}{86.00} & \multicolumn{2}{c}{92.35} \\   
\bottomrule
\end{tabular}
\end{minipage}
\end{table*}

\begin{table*}[!t]
\centering
\begin{minipage}{0.97\linewidth}
\caption{Quantitative comparison on the ImageNet-100 dataset~\cite{ILSVRC15}.}
\label{tbl:result_imagenet100}
\renewcommand{\arraystretch}{1.} 
\centering
\begin{tabular}{ >{\centering}m{0.3\textwidth}| >{\centering}m{0.13\textwidth}| *2{>{\centering}m{0.06\textwidth} >{\centering}m{0.06\textwidth}|} >{\centering}m{0.06\textwidth}  >{\centering\arraybackslash}m{0.06\textwidth} } 
\toprule
\multirow{2}{*}{Method} & \multirow{2}{*}{Backbone} & \multicolumn{2}{c|}{Top-1 Loc} & \multicolumn{2}{c|}{Top-5 Loc} & \multicolumn{2}{c}{GT-known} \\ \cline{3-8}
  &   & $Acc_{avg}$ & $Acc_{last}$ &  $Acc_{avg}$ & $Acc_{last}$ &  $Acc_{avg}$ & $Acc_{last}$ \\
\midrule
AANets~\cite{Liu2021Adaptive} + BAS~\cite{Wu2022BAS}  &  & 29.81  & 27.04  & 40.33  &  38.28 &  48.36 & 49.16  \\
FOSTER~\cite{Wang2022foster} + BAS~\cite{Wu2022BAS}  &   & 36.20 & 38.10 & 44.16 & 47.38  & 48.93 &  50.02 \\
Rebalancing~\cite{Hou2019Rebalancing} + BAS~\cite{Wu2022BAS}  &  \multirow{2}{*}{InceptionV3} & 39.70 & 40.48& 50.27 & 53.66  & 56.06  & 62.04  \\
CCIL~\cite{Mittal2021CCIL} + BAS~\cite{Wu2022BAS}  & \multirow{2}{*}{\cite{inceptionv3}} & 46.33 & 39.84 & 55.25 & 48.38 & 57.33 & 50.62 \\
FDCNet w/o compensation (ours)  &  &  58.12 & 54.92 &  68.05 & 64.54 & 70.21  & 67.63  \\ 
FDCNet (ours)   &     & \textbf{59.52}  & \textbf{56.90} & \textbf{69.75} & \textbf{66.46}  & \textbf{71.48}  & \textbf{69.34} \\ \cline{1-1}\cline{3-8}
Joint training BAS~\cite{Wu2022BAS}  &     & \multicolumn{2}{c|}{68.88} & \multicolumn{2}{c|}{78.74} & \multicolumn{2}{c}{80.14} \\ 
\midrule
AANets~\cite{Liu2021Adaptive} + BAS~\cite{Wu2022BAS} & & 36.79 & 32.12 & 53.60 & 49.98 & 65.71 & 64.60 \\ 
FOSTER~\cite{Wang2022foster} + BAS~\cite{Wu2022BAS} & & 36.60 & 32.20 & 48.20 & 43.96 & 54.25 & 49.84 \\ 
Rebalancing~\cite{Hou2019Rebalancing} + BAS~\cite{Wu2022BAS} &  \multirow{2}{*}{MobileNetV1} & 44.78 & 40.78 & 56.39 & 52.98 & 63.70 & 62.26 \\
CCIL~\cite{Mittal2021CCIL} + BAS~\cite{Wu2022BAS}  & \multirow{2}{*}{\cite{mobilenetv1}} & 46.49 & 41.72 & 58.72 & 54.08 & 61.92 & 58.64 \\ 
FDCNet w/o compensation (ours)  &  & 58.20 & 57.98 & 71.16 & 71.64 & 74.37 & 76.22 \\ 
FDCNet (ours)   &     & \textbf{58.63}  &  \textbf{58.66} & \textbf{71.60} & \textbf{72.40} & \textbf{74.82} & \textbf{76.94} \\ \cline{1-1}\cline{3-8}
Joint training BAS~\cite{Wu2022BAS} &     & \multicolumn{2}{c|}{68.54} & \multicolumn{2}{c|}{81.04} & \multicolumn{2}{c}{83.16} \\ 
\bottomrule
\end{tabular}
\end{minipage}
\end{table*}



\vspace{1mm}
\noindent\textbf{Baselines.}
Since, to the best of our knowledge, this is the first paper investigating class-incremental WSOL, there are no previous results for comparison. Therefore, we also implemented four baseline methods by merging four different class-incremental learning algorithms and a WSOL method. We use the same method, BAS~\cite{Wu2022BAS}, as ours for WSOL and utilize FOSTER~\cite{Wang2022foster}, AANets~\cite{Liu2021Adaptive}, Rebalancing~\cite{Hou2019Rebalancing}, and CCIL~\cite{Mittal2021CCIL} for class-incremental learning. Specifically, we first bring the architecture and losses of BAS~\cite{Wu2022BAS}, and then merge the components and losses for class-incremental learning from one of the methods~\cite{Wang2022foster, Liu2021Adaptive, Hou2019Rebalancing, Mittal2021CCIL}. We use all the losses from both the class-incremental learning methods and the WSOL method and tried our best to conduct fair comparisons.

\subsection{Results} \label{sec:result}
Tables~\ref{tbl:result_cub200} and \ref{tbl:result_imagenet100} show the quantitative comparisons on the CUB-200 dataset~\cite{cub_dataset} and ImageNet-100 dataset~\cite{ILSVRC15}, respectively. The ``Joint training BAS'' rows show the results using the entire training data during whole training process (\ie not incremental learning). Therefore, these results are considered upper bounds. All the methods are trained using two different backbones: MobileNetV1~\cite{mobilenetv1} and InceptionV3~\cite{inceptionv3}. Overall, our proposed strong baseline method (FDCNet w/o compensation) outperforms all other baseline methods in all the metrics and for all the datasets/backbones. Then, the proposed FDCNet achieves further higher accuracies by compensating the effects of feature drifts using the proposed FDC modules. 

In~\tref{tbl:ablation}, we present an ablation study on the losses and the FDC modules to demonstrate their effects on incremental learning. This study was conducted using the ImageNet-100 dataset~\cite{ILSVRC15} and the MobileNetV1 backbone~\cite{mobilenetv1}. The first row shows the results of using only $\calL_{wsol}$ and $\calL_{kd\text{-}cls}$, which are also used in all the following experiments. $\calL_{wsol}$ is adapted from the baseline WSOL network, and $\calL_{kd\text{-}cls}$ is typically used in class-incremental classification.

\begin{table*}[!t]
\centering
\begin{minipage}{0.97\linewidth}
\caption{Ablation study on losses and FDC modules using the ImageNet-100 dataset~\cite{ILSVRC15} and the MobileNetV1 backbone~\cite{mobilenetv1}.}
\label{tbl:ablation}
\centering
\begin{tabular}{ *5{>{\centering}m{0.065\textwidth}}  >{\centering}m{0.065\textwidth}|  *2{>{\centering}m{0.06\textwidth} >{\centering}m{0.06\textwidth}|} >{\centering}m{0.06\textwidth}  >{\centering\arraybackslash}m{0.06\textwidth} } 
\toprule
\multicolumn{6}{c|}{Method} & \multicolumn{2}{c|}{Top-1 Loc} & \multicolumn{2}{c|}{Top-5 Loc} & \multicolumn{2}{c}{GT-known} \\ 
\midrule
$\calL_{wsol}$  &  $\calL_{kd\text{-}cls}$  & $\calL_{kd\text{-}loc}$  & $\calL_{kd\text{-}feat}$ & $g_{dc\text{-}l}$ & $g_{dc\text{-}c}$ & $Acc_{avg}$ & $Acc_{last}$ &  $Acc_{avg}$ & $Acc_{last}$ &  $Acc_{avg}$ & $Acc_{last}$ \\
\midrule
\checkmark  &\checkmark  & &  & & & 57.80 & 57.08 & 71.16 & 70.96 & 74.29 & 75.42 \\ 
\checkmark  &\checkmark  & \checkmark & \checkmark & & &58.20 & 57.98 & 71.16 & 71.64 & 74.37 & 76.22 \\ 
\checkmark  &\checkmark  & \checkmark & \checkmark & \checkmark & & 58.34 &  58.04 & 71.49 & 72.04 & \textbf{74.82} & \textbf{76.94} \\
\checkmark &\checkmark & \checkmark & \checkmark & \checkmark &\checkmark & \textbf{58.63} & \textbf{58.66} & \textbf{71.60} & \textbf{72.40} & \textbf{74.82} & \textbf{76.94} \\
\bottomrule
\end{tabular}
\end{minipage}
\end{table*}

The second row displays the results of additionally using $\calL_{kd\text{-}loc}$ and $\calL_{kd\text{-}feat}$\footnote{$\calL_{kd\text{-}feat}$ includes both $\calL_{kd\text{-}feat\text{-}cls}$ and $\calL_{kd\text{-}feat\text{-}loc}$}. As shown in the second row, the additional losses improve both GT-known and Top-1 Loc accuracies. Since GT-known Loc accuracy evaluates only localization performance independently from classification accuracy, the improvements demonstrate that $\calL_{kd\text{-}loc}$ and $\calL_{kd\text{-}feat}$ are valuable for preserving knowledge for localization.

The third row shows the results of additionally employing $g_{dc\text{-}l}$, which corresponds to using $\calL_{dc\text{-}l}$. The improvements in GT-known Loc accuracies verify that $g_{dc\text{-}l}$ clearly benefits in maintaining knowledge on previously learned classes for localization. Because $g_{dc\text{-}l}$ compensates for the effects of feature drifts on only the localization, the improvements of Top-1 and Top-5 Loc accuracies are caused by the improvements in localization.

The bottom row displays the results of additionally using $g_{dc\text{-}c}$, which corresponds to using $\calL_{dc\text{-}c}$. Since it compensates for the effects of feature drifts on only the classification, GT-known Loc accuracies are the same as the results on the third row. Therefore, the improvements in Top-1 and Top-5 Loc accuracies are caused by the improvements in classification. These gains verify that it is complementary to other components in preserving knowledge of previously learned classes for classification.

\begin{figure}[t]
\centering
    \includegraphics[width=0.37\textwidth]{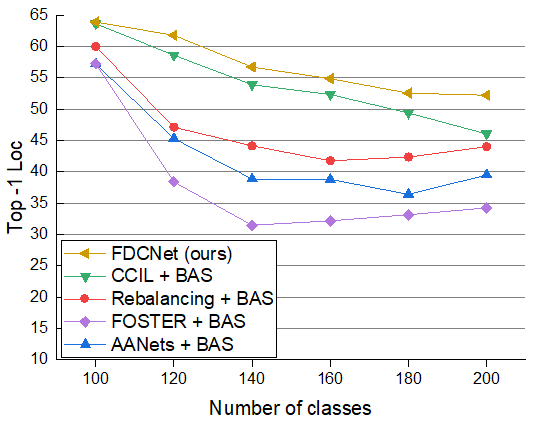}
    \caption{Incremental Top-1 localization accuracy using the MobileNetV1 backbone~\cite{mobilenetv1} on the CUB-200 dataset~\cite{cub_dataset}.} 
\label{fig:incremental_accuracy}
\end{figure}

\fref{fig:incremental_accuracy} shows the incremental Top-1 Loc accuracy achieved using the MobileNetV1 backbone~\cite{mobilenetv1} on the CUB-200 dataset~\cite{cub_dataset}. We compare the results of our proposed method with those of other baseline methods~\cite{Wang2022foster, Liu2021Adaptive, Hou2019Rebalancing, Mittal2021CCIL}. The graph shows $Acc_t$ for $t$ from 1 to $T=6$, which is computed after training on the $t$-th task. While our proposed method achieves a similar accuracy to CCIL~\cite{Mittal2021CCIL} at $t=1$, it achieves higher accuracies for $t>1$. Furthermore, the difference between $Acc_1$ and $Acc_6$ is approximately 10\% for our proposed method, whereas for all other methods, it is more than 15\%. These results demonstrate that our proposed method effectively preserves the knowledge of previously learned classes while learning new object classes.


\begin{figure}[!t]
  \centering
\begin{minipage}{0.2\linewidth}
\centerline{Images}
\end{minipage}
\begin{minipage}{0.17\linewidth}
\centerline{\includegraphics[width=0.8\linewidth,height=0.05\textheight]{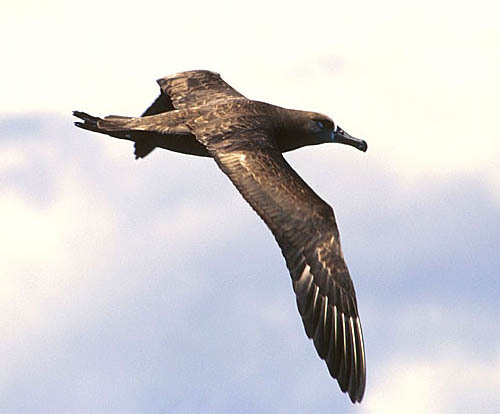}}
\end{minipage}
\begin{minipage}{0.17\linewidth}
\centerline{\includegraphics[width=0.8\linewidth,height=0.05\textheight]{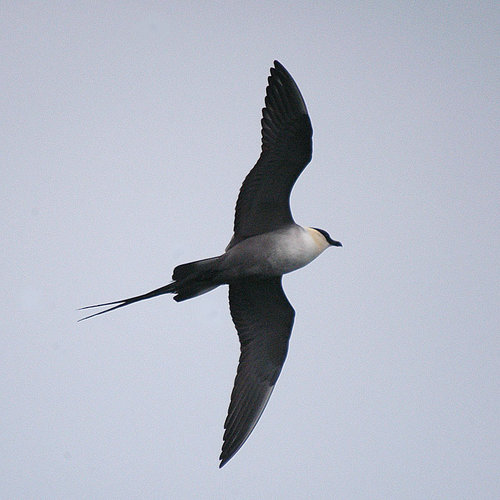}}
\end{minipage}
\begin{minipage}{0.17\linewidth}
\centerline{\includegraphics[width=0.8\linewidth,height=0.05\textheight]{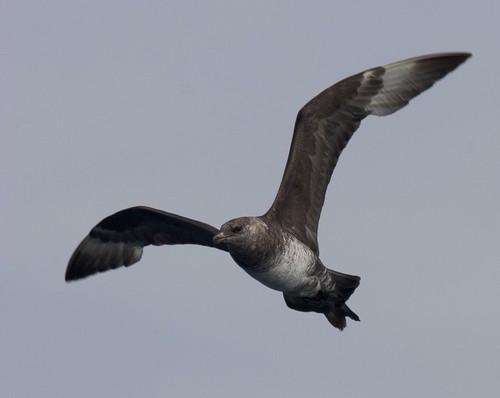}}
\end{minipage}
\begin{minipage}{0.17\linewidth}
\centerline{\includegraphics[width=0.8\linewidth,height=0.05\textheight]{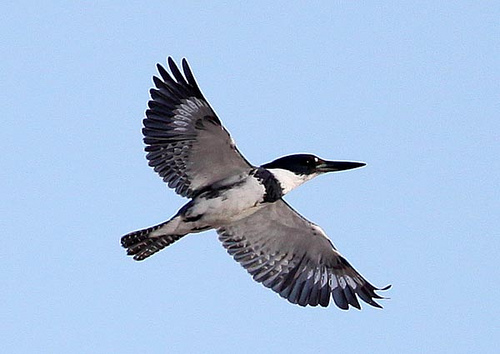}}
\end{minipage}
\\
\vspace{0.01cm}
\begin{minipage}{0.2\linewidth}
\centerline{Rebalancing~\cite{Hou2019Rebalancing}}
\centerline{+BAS~\cite{Wu2022BAS}}
\end{minipage}
\begin{minipage}{0.17\linewidth}
\centerline{\includegraphics[width=0.8\linewidth,height=0.05\textheight]{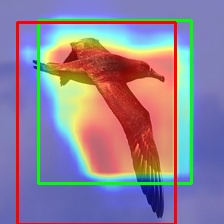}}
\end{minipage}
\begin{minipage}{0.17\linewidth}
\centerline{\includegraphics[width=0.8\linewidth,height=0.05\textheight]{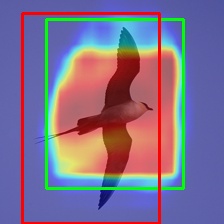}}
\end{minipage}
\begin{minipage}{0.17\linewidth}
\centerline{\includegraphics[width=0.8\linewidth,height=0.05\textheight]{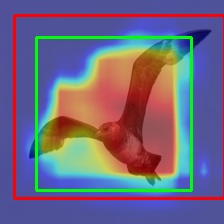}}
\end{minipage}
\begin{minipage}{0.17\linewidth}
\centerline{\includegraphics[width=0.8\linewidth,height=0.05\textheight]{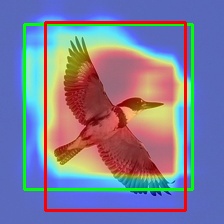}}
\end{minipage}
\\
\vspace{0.01cm}
\begin{minipage}{0.2\linewidth}
\centerline{FDCNet (ours)}
\end{minipage}
\begin{minipage}{0.17\linewidth}
\centerline{\includegraphics[width=0.8\linewidth,height=0.05\textheight]{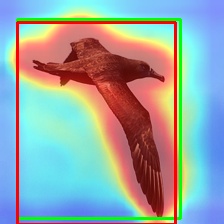}}
\end{minipage}
\begin{minipage}{0.17\linewidth}
\centerline{\includegraphics[width=0.8\linewidth,height=0.05\textheight]{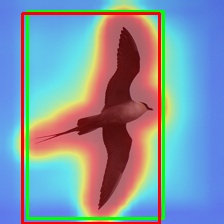}}
\end{minipage}
\begin{minipage}{0.17\linewidth}
\centerline{\includegraphics[width=0.8\linewidth,height=0.05\textheight]{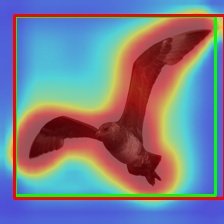}}
\end{minipage}
\begin{minipage}{0.17\linewidth}
\centerline{\includegraphics[width=0.8\linewidth,height=0.05\textheight]{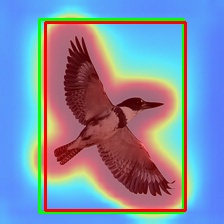}}
\end{minipage}
\vspace{-2mm}
\caption{Qualitative comparison on the CUB-200 dataset~\cite{cub_dataset}.}
\label{fig:result}
\end{figure}


\fref{fig:result} shows qualitative results for localization on the CUB-200 dataset~\cite{cub_dataset}. The results visualize foreground masks of the ground-truth class overlaid on the images. Accordingly, the results correspond to the evaluation setting of GT-known localization. Red and green boxes indicate the ground-truth and predicted results, respectively. The results demonstrate that the FDCNet estimates object location more accurately than the other baseline method.


\section{Conclusion}\label{sec:conclusion}
We introduced a novel task, {\it class-incremental WSOL}, that aims to incrementally learn object localization for novel classes while preserving knowledge to localize previously learned categories. Then, we presented a strong baseline by analyzing the methods for WSOL and class-incremental learning. Then, we proposed the novel \textit{feature drift compensation} (FDC) network to compensate for the effects of feature drifts on both class scores and localization maps to preserve previously learned knowledge while learning new tasks. Lastly, we evaluated the proposed method using the metrics that are extended from those for WSOL to consider class-incremental learning. The results on two public datasets using two backbones demonstrate that our method outperforms other baseline methods. 


\begin{acks}
This work was supported in part by the National Research Foundation of Korea (NRF) grant funded by the Korea government(MSIT) (No.RS-2023-00252434) and by the Korea Evaluation Institute of Industrial Technology (KEIT) grant funded by the Korea government(MOTIE) (No.20018635).
\end{acks}

\bibliographystyle{ACM-Reference-Format}
\balance
\bibliography{sample-base}

\end{document}